%% file: main.tex
\documentclass[11pt,a4paper]{article}
\usepackage[hyperref]{emnlp2017}

\usepackage{times}
\usepackage{color}
\usepackage{amsfonts}
\usepackage{graphicx}
\usepackage{amsmath,amsfonts,amssymb,bm}
\usepackage{booktabs} 
\usepackage{array}
\usepackage{boxedminipage}
\usepackage{paralist}
\usepackage{url}
\usepackage{diagbox}
\usepackage{enumitem}
\usepackage{paralist}
\definecolor{darkspringgreen}{rgb}{0.05, 0.5, 0.06}
\usepackage{arydshln}
\def\emnlppaperid{1301}

\newcommand{\red}[1]{\textcolor{red}{#1}}
\newcommand{\rev}[1]{\textcolor{black}{#1}}

\newcommand{\blue}[1]{\textcolor{blue}{#1}}
\newcommand{\cyan}[1]{\textcolor{darkspringgreen}{#1}}
\newcommand{\sys}{\mbox{\sc NN-Crf}}
\emnlpfinalcopy
\newcolumntype{L}[1]{>{\centering\let\newline\\\arraybackslash\hspace{0pt}}m{#1}}
\newcolumntype{C}[1]{>{\centering\let\newline\\\arraybackslash\hspace{0pt}}m{#1}}
\newcolumntype{R}[1]{>{\centering\let\newline\\\arraybackslash\hspace{0pt}}m{#1}}

\title{Scientific Information Extraction with Semi-supervised Neural Tagging}
\author{
  Yi Luan\quad Mari Ostendorf \quad Hannaneh Hajishirzi \\
 Department of Electrical Engineering, University of Washington\\
   {\{luanyi, ostendor, hannaneh\}@uw.edu}
}

\date{}

\begin{document}

\maketitle

\input{01-Abstract.tex}

\input{02-Introduction.tex}
\input{03-RelatedWork.tex}

\input{04-NeuralStructure.tex}
\input{05-Semi-supervisedLearning.tex}

\input{06-Experiment.tex}

\input{07-Analysis.tex}

\input{08-Conclusion.tex}
\input{09-Acknowledgement.tex}

\bibliography{references}
\bibliographystyle{emnlp_natbib}

\end{document}

%% file: 01-Abstract.tex
\begin{abstract}

This paper addresses the problem of extracting  keyphrases from scientific articles and categorizing them as corresponding to a task, process, or material. We cast the problem as sequence tagging and introduce  semi-supervised methods to a neural tagging model, which builds on recent advances in named entity recognition. Since annotated training data is scarce in this domain, we introduce a graph-based semi-supervised algorithm together  with a data selection scheme to leverage unannotated articles. Both inductive and transductive semi-supervised learning strategies outperform state-of-the-art information extraction performance on the 2017 SemEval Task 10 ScienceIE task. 

\end{abstract}

%% file: 02-Introduction.tex
\section{Introduction}

 As a research community grows, more and more papers are published each year.  As a result there is increasing demand for improved methods for finding relevant papers and automatically understanding the key ideas in those papers. However, due to the large variety of domains and extremely limited annotated resources, there has been relatively little work on scientific information extraction. Previous research has focused on unsupervised approaches such as bootstrapping~\cite{gupta2011analyzing,tsai2013concept}, where hand-designed templates are used to extract scientific keyphrases, and more templates are added through bootstrapping. 

 Very recently a new challenge on Scientific Information Extraction (ScienceIE) \cite{scienceIE}\footnote{SemEval (Task 10)\url{https://scienceie.github.io/index.html}} provides a dataset consisting of 500 scientific paragraphs with keyphrase annotations  for three categories: \textsc{Task}, \textsc{Process}, \textsc{Material} across three scientific domains, Computer Science (CS), Material Science (MS), and Physics (Phy), as in Figure~\ref{fig:ppl}. This dataset enables the use of more advanced approaches such as neural network (NN) models. To that end, we cast the keyphrase extraction task as a sequence tagging problem, and build on recent progress in another information extraction task:  Named Entity Recognition  (NER)~\cite{lample2016neural,peng2015named}. 
 Like named entities, keyphrases can be identified by their linguistic context, e.g. researchers "use" methods. In addition, keyphrases can be associated with different categories in different contexts. For example, `semantic parsing' can be labeled as a \textsc{task} in one article and as a \textsc{process} in another. Scientific keyphrases differ in that they can include both noun phrases and verb phrases and in that non-standard ``words'' (equations, chemical compounds, references) can provide important cues.

\begin{figure}
\centering

\begin{footnotesize}
\begin{tabular}{p{7.2cm}}
\toprule
{\it \textbf{Computer Science:}} \\ This paper addresses the task of \textbf{[named entity recognition]$\blue{\mathrm{_{\bm{Task}}}}$}, using  \textbf{[conditional random fields]$\blue{\mathrm{_{\bm{Process}}}}$}. Our method  is evlauated on the \textbf{[ConLL NER Corpus]$\blue{\mathrm{_{\bm{Material}}}}$}. 
\\
\midrule
{\it \textbf{Physics:}} \\ \textbf{[Local field effects] $\blue{\mathrm{_{\bm{Process}}}}$} on spontaneous emission rates within \textbf{[nanostructure photonics material]$\blue{\mathrm{_{\bm{Material}}}}$} for example are familiar, and have been well used.\\
\midrule
{\it \textbf{Material Science:}} \\ The \textbf{[Kelvin probe force microscopy technique] $\blue{\mathrm{_{\bm{Process}}}}$} allows \textbf{[detection of local EWF]$\blue{\mathrm{_{\bm{Task}}}}$} between an  \textbf{[atomic force micorscopy]$\blue{\mathrm{_{\bm{Material}}}}$} and  \textbf{[metal surface]$\blue{\mathrm{_{\bm{Material}}}}$}.\\
\bottomrule
\end{tabular}
\end{footnotesize}
\footnotesize{\caption{Annotated ScienceIE examples.}}
\label{fig:ppl}
\end{figure}




Since the scale of the  data is still small for supervised training of neural systems, 
we introduce  semi-supervised methods to the neural tagging model in order to take advantage of the large quantity of unlabeled scientific articles. This is particularly important because of the differences in keyphrases across domains.
Our semi-supervised learning algorithm  uses a graph-based label propagation scheme to estimate the posterior probabilities of unlabeled data. It additionally extends the  training objective to leverage the confidence of the estimated posteriors.  
The new  training  treats low confidence tokens as missing labels and computes the sentence-level score by marginalizing over them. 

Our experiments show that our neural tagging model achieves state-of-the-art results in the SemEval Science IE task. We further show that both inductive and transductive semi-supervised strategies significantly improve the performance.  Finally, we provide in-depth analysis of domain differences as well as analysis of failure cases. 

The key contributions of our work include: i)~achieving state of the art in scientific information extraction SEMEVAL Task 10 by extending recent advances in neural tagging models; ii)~introducing a semi-supervised learning algorithm that uses graph-based label propagation and confidence-aware data selection,  iii)~exploring different alternatives for taking advantage of large, multi-domain unannotated data including both unsupervised embedding initialization and semi-supervised model training.




%% file: 03-RelatedWork.tex
\section{Related Work}
There has been growing interest in research on automatic methods to help researchers search and extract useful information from scientific literature. Past research has addressed citation sentiment~\cite{athar2012detection,athar2012context}, citation networks~\cite{kas2011structures,GBOR16.870, sim2012discovering,do2013extracting,jaidka2014computational}, summarization~\cite{abu2011coherent} and some analysis of research community~\cite{vogel2012he,anderson2012towards,luan2012performance,luan2014relating,levow2014recognition}. However, due to scarce hand-annotated data resources, previous work on information extraction (IE) for scientific literature is very limited.  \citet{gupta2011analyzing} first proposed a task that defines scientific terms for 474 abstracts from the ACL anthologhy~\cite{bird2008acl} into three aspects: \textit{domain}, \textit{technique} and \textit{focus} and apply template-based bootstrapping to tackle the problem. Based on this study, \citet{tsai2013concept} improve the performance by introducing hand-designed features from NER \cite{collins1999unsupervised} to the bootstrapping framework.  \citet{qasemizade} compile a dataset of scientific terms into 7 fine-grained categories for 171 abstracts of ACL anothology.  Similar to our work, very recently \citet{augenstein2017multi} also evaluated on ScienceIE dataset, but use multi-task learning to improve the performance of a supervised neural approach. Instead, we introduce a semi-supervised neural tagging approach that leverages unlabeled data. 

Neural tagging models have been recently introduced to tagging problems such as NER. For example,  \citet{collobert2011natural} use a CNN over a sequence of word embeddings and apply a CRF layer on top.  \citet{huang2015bidirectional} use hand-crafted features with LSTMs to improve performance. There is currently great interest in using character-based embeddings in neural models.  \cite{chiu2015named,lample2016neural,ballesteros2015improved,ma2016end}. Our approach also takes advantage of neural tagging models and character-based embeddings for IE in scientific articles. 

\rev{ Previous work on semi-supervised learning for neural models has mainly focused on transfer learning \cite{dai2015semi,luan2014semi,harsham2015driver} or initializing the model with pre-trained word embeddings \cite{mikolov2013efficient,pennington2014glove,levy2014dependency,luan2016lstm,luan2015efficient,luan2016multiplicative}. In our work, we use pre-training but also use more powerful methods including graph-based semi-supervision \cite{subramanya2011semi,liu2013graph,liu2015acoustic,liu2016graph,liu2016novel} and a method for leveraging partially labeled data \cite{kim2015weakly}. We show that the combination of these techniques gives better results than any one alone.}



%% file: 04-NeuralStructure.tex
\section{Problem Definition and Data}
\label{sec:IOB}

The purpose of this work is to extract phrases that can answer questions that researchers usually face when reading a paper: What \textsc{Task} has the paper addressed? What \textsc{Process} or method has the paper used or compared to? What \textsc{Materials} has the paper utilized in experiments? While these fundamental concepts are important in a wide variety of scientific disciplines, the terms that are used in specific disciplines can be substantially different. For example, \textsc{Materials} in computer science might be a text corpus, while they would be physical materials in physics or materials science. 
\vspace{-.2cm}
\paragraph{Data} We use the SemEval 2017 Task 10 ScienceIE dataset.  Fig.~\ref{fig:ppl} provides examples that illustrate the variation in domains, but also show that there are common cues such as ``the task of", ``using", ``technique," etc.
A challenge with this dataset is that the size of the training data is very small. It is built from ScienceDirect open access publications and consists of 500 journal articles, but only one paragraph of each article is manually labeled. Therefore, we use a large amount of external data to leverage the continuous-space representation of language in neural network model. We explore the effect of pre-training word embedding with two different external resources: i)~a data set of Wikipedia articles as a general English resource, and ii)~a data set of 50k Computer Science papers from ACM.\footnote{Due to the difficulty of data collection, experiments with external data from the other two domains is left to future work.}
\vspace{-.2cm}
\paragraph{Tagging Problem Formulation}
The task requires detecting the exact span of a keyphrase. In order to be able to distinguish spans of two consecutive keyphrases of the same type, we assign labels to every word in a sentence, indicating position in the phrase and the type of phrase. We formulate the problem as an IOBES (Inside, Outside, Beginning, End and Singleton) tagging problem where every  token  is  labeled either as:  B, if it is at the beginning  of  a  keyphrase;  E, if it ends the phrase; I, if  it  is  inside
a  keyphrase  but  not  the  first  or last token; S, if it is a single-word keyphrase; or  O,  otherwise. For example, ``named entity recognition" in first sentence of Fig.~\ref{fig:ppl} is labeled as ``\textit{B-Task} \textit{I-task} \textit{E-task}".


\section{Neural Architecture Model}
\label{sec:NN}
We introduce an end-to-end model to categorize scientific keyphrases, building on a neural named entity recognition model~\cite{lample2016neural} and adding a feature-based embedding. 

\subsection{Model}
\label{sec:model}
We develop a 3-layer hierarchical neural model  to tag  tokens of the documents (details of the tokenization is in Sec.~\ref{sec:exp_setup}). 
    (1) The {token representation layer} concatenates   three components for each token: a bi-directional character-based embedding, a word embedding, and an embedding associated with orthographic and part-of-speech features. (2) The token LSTM layer uses a bidirectional LSTM to incorporate contextual cues from surrounding tokens to derive intermediate token embeddings.    
     (3) The CRF tagging layer models token-level tagging decisions jointly using a CRF objective function to incorporate dependencies between tags. 
\vspace{-.2cm}
\paragraph{Character-Based Embedding.}
The embedding for a token is derived from its characters as the concatenation
of  forward  and  backward  representations  from
a bidirectional LSTM. The
character lookup table is initialized at random. The advantage of building a character-based embedding layer is that it can handle out-of-vocabulary words and equations, which are frequent in this data, all of which are mapped to ``UNK" tokens in the Word Embedding Layer. 
\vspace{-.2cm}
\paragraph{Word Embedding.} Words from a fixed vocabulary (plus the unknown word token) are mapped to a vector space, initialized using Word2vec pre-training with different combinations of corpora. 

\vspace{-.2cm}\paragraph{Feature Embedding.}
We map  features to a vector space:  capitalization (all capital, first capital, all lower, any capital but first letter) and Part-of-Speech tags.\footnote{Dependency features were investigated but did not lead to performance gains.} We randomly initialize feature vectors and train them together as other parameters.

\vspace{-.2cm}
\paragraph{Token LSTM Layer}
We apply a bidirectional LSTM at the token level taking the concatenated character-word-feature embedding as input. 
The token representation obtained by stacking the forward and backward LSTM hidden states is passed as input to a linear layer that project the dimension to the size of  label type space and is used as input to CRF layer. 
\vspace{-.2cm}
\paragraph{CRF Layer} 
Keyphrase categorization is a task where there is strong dependencies across output labels (e.g., I-TASK cannot follow B-Process). Therefore, instead of making independent tagging decisions for each output, we  model them jointly using conditional random field \cite{lafferty2001conditional}.
For an input sentence $\bm{x}= (x_1,x_2,x_3,\dots,x_n)$, we consider $P$ to be the matrix of scores output by
the bidirectional LSTM network. $P$ is of size $n \times m$,
where $n$ is the number of tokens in a sentence, and $m$ is the number of distinct tags.  $P_{t,i}$ corresponds to the score of the $i$-th tag of the $t$-th word
in a sentence.
We use a first-order Markov Model and define a transition matrix $\bm{T}$ where $\bm{T}_{i,j}$ represents the score from tag $i$ to tag $j$. We also add $y_0$ and $y_n$ as the \textit{start} and \textit{end} tags of a sentence. Therefore $\bm{T}$ becomes a square matrix of dimension $m+2$. 

\noindent Given one possible output $\bm{y}$, and neural network parameters $\theta$ we define the score as
\begin{equation}
\label{eq:fi}
\phi(\bm{y};\bm{x},\theta) = \sum_{t=0}^n T_{y_t,y_{t+1}} + \sum_{t=1}^n P_{t,y_t}
\end{equation}
The probability of sequence $\bm{y}$ is obtained by applying a softmax over all possible tag sequences
\begin{equation}
\label{eq:score}
p_\theta(\bm{y}|\bm{x}) =  \frac{\exp(\phi(\bm{y};\bm{x},\theta))}{\sum_{\bm{y}'\in Y} \exp(\phi(\bm{y}';\bm{x},\theta))}
\end{equation}
\noindent where $Y$ denotes all possible tag sequences. The normalization term is efficiently computed  using the forward algorithm.

\paragraph{Supervised Training}  During training, we maximize the log-probability $\mathcal{L}(\bm{Y};\bm{X},\theta)$ of the correct tag sequence given the corpus $\{\bm{X},\bm{Y}\}$.  Back-propagation is done based on a gradient computed using sentence-level scores. 



%% file: 05-Semi-supervisedLearning.tex
\section{Semi-supervised Learning}
We develop a semi-supervised algorithm that extends self-training by estimating the labels of unlabeled data and then using those labels for re-training. Specifically, we use a  graph-based algorithm to estimate the posterior probabilities of unlabeled data 
and develop a new  CRF training to take the uncertainty of the estimated labels into account while optimizing the objective function.

\begin{figure}[t]
\centering
\includegraphics[width=8cm]{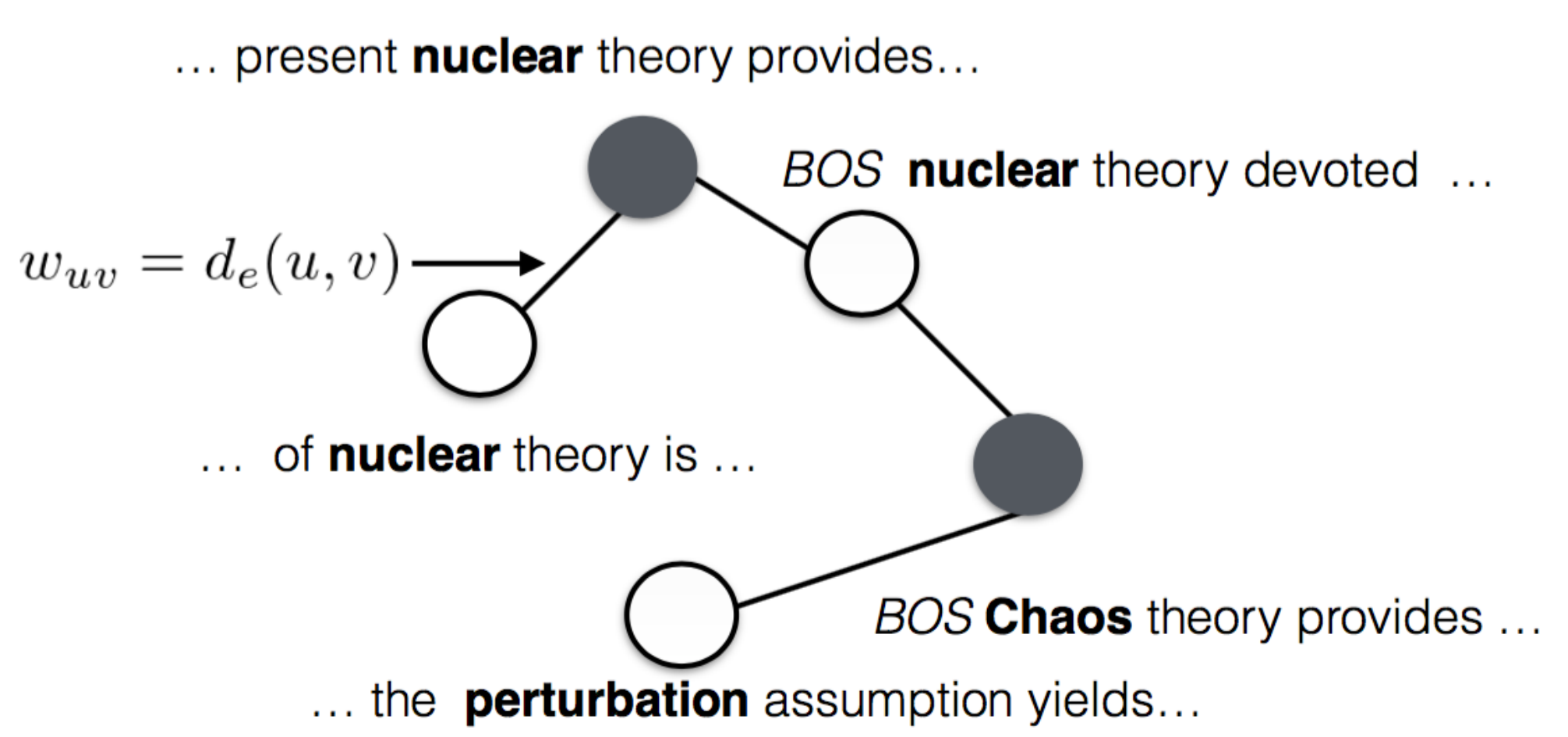}           
\caption{\small{Label propagation. Gray nodes indicates labeled data while white nodes are unlabeled. Bold font word indicates the current token. 
The assumption is if two instances are similar according to the graph, the output labels should be similar.
}}
\label{fig:graph}
\end{figure}

\subsection{Graph-based Posterior Estimates}
\label{sec:graph}
Our semi-supervised algorithm uses the following steps to estimate the posterior. It first constructs a graph of tokens based on their semantic similarity, then uses the CRF marginal as a regularization term to do label propagation on the graph. The smoothed posterior is then used to either interpolate with the CRF marginal or as an additional feature to the neural network.

\vspace{-.2cm}
\paragraph{Graph Construction} Vertices in the graph correspond to tokens, and edges are distance between token features which capture semantic similarity. The total size of the graph is equal to the number of tokens in both labeled data $V_l$ and unlabeled data $V_u$. The tokens are modelled with a concatenation of pre-trained word embeddings (with dimension $d$) of 5-gram centered by the current token, the word embedding of the closest verb,  and a set of discrete features including part-of-speech tags and capitalization (43 and 4 dimension one-hot features). The resulting feature vector with dimension of $5d+d+43+4$ is  then projected down to 100 dimensions using PCA.   We define the weight $w_{uv}$ of the edge between nodes   $u$ and $v$ as follows: $w_{uv} = d_e(u,v) \text{ if } v \in \mathcal{K}(u) \text{ or } u\in \mathcal{K}(v)$,   where $\mathcal{K}(u)$ is the set of \textit{k}-nearest neighbors of $u$ and $d_e(u,v)$ is the Euclidean distance between any two nodes $u$ and $v$ in the graph. An example of our graph is in Fig.~\ref{fig:graph}. 


For every node $i$ in the graph, we compute the marginal probabilities $\{ \bm{q}_i\}$ using the forward-backward algorithm. Let $\theta^i$ represent the estimate of the CRF parameters after the $n$-th iteration, we compute  the marginal probabilities 
$\tilde{\bm{p}}_{(j,t)} = p(y_t^{j}|\bm{x};\theta^i)$ over IOBES tags for every token position $t$ in sentence $j$ in labeled and unlabeled data. 

\vspace{-.2cm}
\paragraph{Label Propagation} 
We use prior-regularized measure propagation \cite{liu2014graph,subramanya2011semi} to propagate labels from the annotated data to their neighbors in the graph. The algorithm aims for the label distribution between neighboring nodes to be as similar to each other as possible by optimizing an objective function that
minimizes the Kullback-Leibler distances between: i) the empirical distribution $\bm{r}_u$  of labeled data and the predicted label distribution $\bm{q}_u$ for all labeled nodes in the graph; ii) the distributions $\bm{q}_u$ and $\bm{q}_v$ for all nodes $u$ in the graph and their neighbors $v$; iii) the distributions $\bm{q}_u$ and the CRF marginals $\tilde{\bm{p}}_u$ for all nodes. 
The third term regularizes the predicted distribution toward the CRF prediction if the node is not connected to a labeled vertex, ensuring the algorithm performs at least as well as standard self-training. 

\vspace{-.2cm}
\paragraph{Posterior Estimates}
We develop two strategies to estimate the new posteriors $\hat{p}(y_t|\bm{x};\theta)$, which can then be used in the CRF training.   

The first strategy (called \textsc{GraphInterp}) is the commonly used approach~\cite{subramanya2010efficient,aliannejadi2014graph}  that interpolates the smoothed posterior  $\{\bm{q}\}$ with CRF marginals $p$:
\begin{equation}
\label{eq:inductive}
\hat{p}(y_t|\bm{x};\theta) = \alpha p(y_t|\bm{x};\theta) + (1-\alpha) q(y)
\end{equation}
\noindent where $\alpha$ is a mixing coefficient. 

A second strategy introduced here (called \textsc{GraphFeat}) uses the smoothed posterior $\{\bm{q}\}$ as  features and learns it with other parameters in the neural network. 
Given a sentence $\{x_1,\dots,x_{n}\}$, let $Q=\{\bm{q}_1,\dots,\bm{q}_n\}$ be the predicted label distribution from the graph. We then use $Q$ as a feature input to neural network as 
 $\tilde{P} = P + M Q$
 where $P$ is the $n \times m$ matrix output by  the  bidirectional  LSTM network as in Eq. \ref{eq:fi}, and $M$ is $m\times m$ matrix and is learned together with other parameters of neural network.
 We modify Eq. \ref{eq:fi} by replacing ${P}_{t,y_t}$ with  $\tilde{P}_{t,y_t}$.
%
Note that \textsc{GraphFeat} can only be done in a transductive way  since it requires output $Q$ from the graph at test time.

\subsection{CRF training with Uncertain Labels}
\label{sec:ulm}

\begin{figure}[t]
\footnotesize
\centering
\includegraphics[width=8cm]{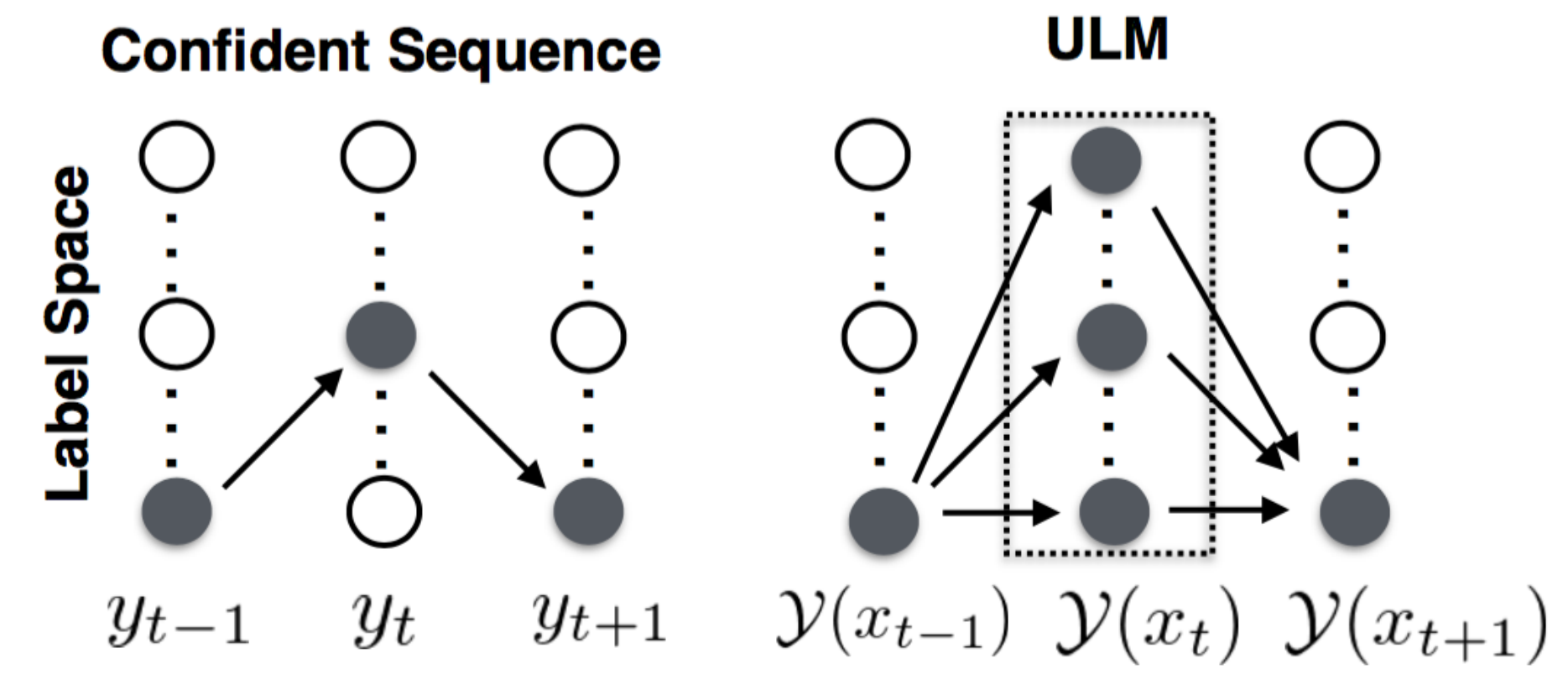}           
\caption{\small{Lattice representation of ULM. Dashed box is the uncertain token which is going to  be marginalized over. \rev{Arrows and grey nodes are paths to be summed over during training. When all tokens are confident, the score of only one path is calculated.}}}
\label{fig:ULM}
\end{figure}
A standard approach to self-training is to make hard decisions for labeling tokens based on the estimated posteriors and retrain the model.  However, the  
estimated posteriors in our task are noisy due to the difficulty and variety of the ScienceIE task. 
Instead,  we extend the CRF training to leverage the confidence of the estimated posteriors.  
The new CRF training (called Uncertain Label Marginalizing (ULM)) treats low confidence tokens as missing labels and computes the sentence-level score by marginalizing over them. A similar idea has been previously used in treating partially labeled data~\cite{kim2015weakly}.

Specifically, given a sentence $\bm{x}$ we define a constrained \textit{lattice} $\mathcal{Y}(\bm{x})$, where at each position $t$ the allowed label types $\mathcal{Y}(x_t)$ are:
\begin{equation}
\mathcal{Y}(x_t) = 
\begin{cases}
\{y_t\}, \qquad \text{if} \quad p(y_t|\bm{x};\theta) > \eta\\
\text{All label types}, \quad \text{otherwise}
\end{cases}
\label{eq:ulm}
\end{equation}
where $\eta$ is the confidence threshold, $y_t$ is the prediction of posterior decoding and $p(y_t|\bm{x};\theta)$ is its CRF token marginal. The new neural network parameters $\theta$ are estimated by maximizing the log-likelihood of $p_\theta(\mathcal{Y}(\bm{x}^{k})|\bm{x}^{k})$ for every input sentence $\bm{x}^{k}$, where
\begin{equation*}
\label{eq:latscore1}
p_\theta(\mathcal{Y}(\bm{x}^{k})|\bm{x}^k) =  \frac{\sum_{\bm{y}^{k}\in \mathcal{Y}(\bm{x}^{k})}\exp(\phi(\bm{y}^k;\bm{x}^k,\theta))}{\sum_{\bm{y}'\in Y} \exp(\phi(\bm{y}';\bm{x},\theta))}
\end{equation*}
where $\bm{y}^{k}$ is an instance  sequence of lattice $\mathcal{Y}(\bm{x})$,  
and $k$ is the sentence index in the training set. 
Extreme cases are when all tokens are uncertain then the likelihood would be equal to 1, when all tokens of a sequence are confident, it would be equal to Eq. \ref{eq:score} where only one possible sequence, as in Fig.~\ref{fig:ULM}. 
\vspace{-.2cm}
\paragraph{Inductive and Transductive Learning} 
The semi-supervised training process is summarized as follow: It first computes marginals over the unlabeled data given  a set of CRF parameters. 
It then uses the marginals as a regularization term for label propagation. The smoothed posteriors from the graph are then interpolated with the CRF marginal in \textsc{GraphInterp} or used as an additional feature in \textsc{GraphFeat}. It then uses the estimated labels for  the unlabeled data combined with the labeled data to retrain the CRF using either the hard decision CRF training objective as Eq.~\ref{eq:score}  or the ULM data selection objective. 

In the inductive setting, we only use the unlabeled data from the development set for the semi-supervision. In the transductive setting we also use the unlabeled data of the test set to construct the graph. In both cases, the parameters are tuned only on the dev set. 

%% file: 06-Experiment.tex
\begin{table*}
  \centering
 {\footnotesize
  \begin{tabular}{llll}
    \toprule
  Span Level  & Classification (dev) & Classification (test)  & Identification \\
  \midrule
    Gupta et.al.(unsupervised)& - & 9.8 & 6.4\\
    Tsai et.al. (unsupervised) & - & 11.9 & 8.0\\  \hdashline
        \textsc{MultiTask} & 45.5 & - & -\\
    Best Non-Neural SemEval$^+$ & - & 38 & 51 \\
    Best Neural SemEval$^+$ & - & 44 & 56\\
    \sys(supervised) & 48.1 & 40.2 & 52.1\\
    \sys(semi)  & 51.9 & 45.3 & 56.9\\ \hdashline
    \sys(semi)$^*$ & \textbf{52.1} & \textbf{46.6} & \textbf{57.6}\\
    \bottomrule
  \end{tabular}}
  \caption{ \small{Overall span-level F1 results for keyphrase identification (SemEval Subtask A) and classification (SemEval Subtask B). $^*$ indicates tranductive setting. $^+$ indicates not documented as either transductive or inductive. - indicates score not reported or not applied.}}
  \label{tab:best_span}
\end{table*}

\section{Experimental Setup}
\label{sec:exp_setup}

\vspace{-.2cm}
\paragraph{Data} The SemEval ScienceIE (\textsc{SE}) corpus consists of 500 journal articles; one paragraph of each article is randomly selected and annotated. The complete unlabeled articles and their metadata are provided together with the labeled data. The training data consists of 350 documents; 50 are kept for development and 100 for testing. The 500 articles come from 82 different journals evenly distributed in three domains. We manually labeled 82 journal names in the dataset into the three domains and do analysis based on the domain partitions.  The 500 full articles contains 2M words and is 30 times the size of the annotated data.

Additionally, we use two external resources for pretraining word embeddings: i)~\textsc{wiki}, as for  Wikipedia articles, specifically a full Wikipedia dump from 2012 containing 46M words, and ii)~\textsc{ACM}, a collection of CS papers, containing 108M words. 

\vspace{-.2cm}
\paragraph{Comparisons} We compare our system with two template matching baselines and the state-of-the-art on the SemEval Science IE task. The first baseline~\cite{gupta2011analyzing} is  an unsupervised method to extract keyphrases by initially using  seed patterns in a dependency tree, and then adding to seed patterns through bootstrapping.  
The second baseline \cite{tsai2013concept} improves the work of \newcite{gupta2011analyzing} by adding Named Entity Features and use different set of seed patterns.  

\begin{table}
  \centering
 {\footnotesize
   \begin{tabular}{l|lll}
    \toprule
     Model & P & R & F1 \\    
    \midrule
    \sys(supervised) & 46.2 & 48.2 &47.2\\
     \midrule
     \ \ \ No features & 44.2& 46.1 & 45.1\\
     \ \ \ No bi-LSTM & 45.2 & 44.7 &44.9\\
     \ \ \ No CRF & 36.7 & 38.2 & 37.4\\
     \ \ \ No char & 45.7 & 46.2 & 45.9\\
    \bottomrule
  \end{tabular}}
  \caption{ \small{Ablation study showing impact of neural network configurations of our \sys(supervised) model on the dev set.} }
  \label{tab:ablation}
\end{table}

\vspace{-.2cm}
\paragraph{Implementation details} 

All parameters are tuned on the dev set performance, the best parameters are selected and fixed for model switching and semi-supervised systems.  The word embedding dimension is 250; the token-level hidden dimension is 100; the character-level hidden dimension is 25; and the optimization algorithm is SGD with a learning rate of 0.05. For building the graph, the best pre-trained embeddings for the supervised system (Sec.~\ref{sec:exp_emb}) are used in each domain. Two special tokens \textit{BOS} and \textit{EOS} are added when pre-training, indicating the begin and end of a sentence. The number of the graph vertices is 2M in tranductive setting and 1.4M in inductive setting. The ULM parameter $\eta$ in Eq. \ref{eq:ulm} is tuned from 0.1 to 0.9, the best $\eta$ is  0.4. The best parameters of label propagation are $\mu=10^{-6}$ and $\nu=10^{-5}$. The interpolation parameter $\alpha$ in Eq.~\ref{eq:inductive} is tuned from 0.1 to 0.9, the best $\alpha$ is  0.3. \rev{We do iteration of semi-supervised learning until we obtain the best result on the dev set, which is mostly achieved in the second round.}

We use Stanford CoreNLP~\cite{manning2014stanford} tokenizer to tokenize words. The tokenizer is augmented with a few hand-designed rules to handle equations (e.g. ``fs(B,t)=Spel(t)S" is a single token) and other non-standard word phenomena (Cu40Zn, 20MW/m2) in scientific literature. We use Approximate Nearest Neighbor Searching (ANN)\footnote{\url{https://www.cs.umd.edu/~mount/ANN/}} to calculate the \textit{k}-nearest neighbors. For all experiments in this paper, $k=10$.

\vspace{-.2cm}
\paragraph{Setup} We evaluate our system in both inductive and transductive settings. The systems with a $^*$ superscript in the table are transductive.  The inductive setting uses 400 full articles in ScienceIE training and dev sets, while the transductive setting uses 500 full articles including the test set. In both settings parameters are tuned over the dev set.

\section{Experimental Results}
We  evaluate our \sys\ model in both supervised and semi-supervised settings. We also perform ablations and try different variants to best understand our model. 

\subsection{Best Case System Performance}

 Table \ref{tab:best_span} reports the results of our neural sequence tagging model \sys\ in both  supervised  and semi-supervised learning (ULM and graph-based),  and compares them with the  baselines and the  state-of-the-art (best SemEval System~\cite{scienceIE}).
 
 \rev{\citet{augenstein2017multi} use a multi-task learning strategy to improve the performance of supervised keyphrase classification, but they only report dev set performance on SemEval Task 10, we also include their result here and refer it as \textsc{MultiTask}.} We report results for both span identification (SemEval SubTask A) and span classification into \textsc{Task}, \textsc{Process} and \textsc{Material} (SemEval Subtask B).\footnote{The evaluation script is provided by the challenge, with a modification to report 3 decimal precision results.}
 
 The results show that our neural sequence tagging models significantly outperforms the state of the art and both baselines.  It confirms that our neural tagging model outperforms other non-neural and neural models for the SemEval ScienceIE challenge\footnote{Best SemEval Numbers from https://scienceie.github.io/}. 
  It further shows that our system achieves significant boost from semi-supervised learning using unlabeled data. Table~\ref{tab:summary} shows the detailed analysis of the system across different categories. 
\subsection{Supervised Learning}
\label{sec:exp_emb}

\paragraph{Impact of Neural Model Components}
Table \ref{tab:ablation} provides the results of an ablation study on the dev set showing the impact of different components of our \sys\ on the Scientific IE task. For the basic model, the word embeddings are initialized by word2vec trained on the 350 full journal articles in the SE training set together with Wikipedia and ScienceIE data. The feature layer, character layer, and bi-LSTM word layers all improves the performance. Moreover, we observe a large improvement  (20.6\% relative) in the scientific IE task by adding the CRF layer.  

\begin{table}
  \centering
 {\footnotesize
   \begin{tabular}{l|lll|lll}
    \toprule
    & \multicolumn{3}{c}{Dev} & \multicolumn{3}{c}{Test}  \\  
     Initialization & MS & Phy & CS  & MS & Phy & CS \\
    \midrule
    SE & 49.4 & 39.4 & 45.0 &42.9 & 33.0 & 30.5\\
    +wiki & \textbf{52.9} & \textbf{40.5} & 47.9 &\textbf{46.1} & \textbf{39.2} & 31.0 \\
    +ACM & 50.3 & 39.8 & \textbf{49.5}  & 42.2 & 37.8 & 34.2  \\
    +wiki+ACM & 50.5 & 40.3 & 48.9 & 43.1 & 37.9 & \textbf{34.4} \\

    \bottomrule
  \end{tabular}}
  \caption{ \small{F1 score on the dev and test sets for using different sources of data for pretraining. }} 
  \label{tab:embedding}
\end{table}

\vspace{-.2cm}
\paragraph{Initialization} Table \ref{tab:embedding} reports our \sys\ performance  when pretrained on different domains.  We explore different word embedding  pre-training with ScienceIE training set alone (SE), and adding other external resources including Wikipedia (wiki) and  Computer Science articles (ACM). All alternatives use word2vec. 
Compared with using SE alone, introduction of all external data sources improve performance. Moreover, we observe that with the introduction of the ACM dataset, the performance on the CS domain is increased significantly in both the dev and test sets. Adding Wikipedia data benefits all three domains, with more significant improvement on the MS and Physics domains.

Based on these observations, we select the best model on each domain according to the dev set  and use the combined result as our best suprevised system (called \sys(supervised)). The F1 score improves from 39.4 to 40.2 when applying model switching strategy.
The best  model on the dev set  is used for each domain: for MS and physics domain, we pretrain word embeddings with the SE and Wiki, and for the CS domain, we pretrain with the SE and ACM. 


\subsection{Semi-Supervision Learning}
\label{sec:exp-ssl}
Table~\ref{tab:ssl} reports the results of the semi-supervised learning algorithms in different settings. In particular we ablate incorporating the graph-based methods of computing the posterior and CRF training (ULM vs. hard decision). The table shows incorporating graph-based methods for computing posterior and ULM for CRF training outperforms their counterparts. 
\begin{table}
  \centering
 {\footnotesize
   \begin{tabular}{ll|ll}
    \toprule
     Posterior & Training & Dev & Test\\
     \midrule
     - & -& 50.2 & 42.9 \\
     - &ULM & 51.3 & 44.4\\
     \textsc{GraphInterp}&- & 50.9 & 43.3\\
\textsc{GraphInterp} & ULM & \textbf{51.9} & \textbf{45.3} \\ \hline
               \textsc{GraphInterp}* &-& 50.7 & 44.0\\
\textsc{GraphInterp}* &ULM& 51.8 & 45.7\\
     \textsc{GraphFeat}* &- & 51.4 & 44.9 \\ 
\textsc{GraphFeat}* &ULM & \textbf{52.1} & \textbf{46.6} \\
    \bottomrule
  \end{tabular}}
  \caption{ \small{F1 scores of semi-supervised Learning approaches; * shows transductive models. }}
  \label{tab:ssl}
\end{table}

For computing the posterior, we explore two different strategies of the graph-based methods: i) \textsc{GraphInterp} that interpolates the smoothed posterior from label propagation with CRF marginals; For inductive setting, \textsc{GraphInterp} only uses un-annotated data from the dev set and uses the best model for decoding at test time. For transductive setting, \textsc{GraphInterp}$^*$ uses un-annoated data from test set to build the graph as well, and  tune the parameters on the dev set.  ii) \textsc{GraphFeat} uses the smoothed posterior from label propagation as additional feature to neural network and only has transductive setting.

\begin{table}
  \centering
 {\small
  \begin{tabular}{l|llll}
    \toprule
   Span Level & T & P & M & K  \\
  \midrule
    Best SemEval& 19 & 44 & 48 & 55\\
   supervised & 13.3 & 40.5 & 43.7 & 52.1\\
   \textsc{ULM+GraphInterp} & 17.0 & 45.4 & 49.4 & 56.9\\ 
   \textsc{ULM+GraphFeat}* & 17.2 & 46.5 & 50.7 & 57.6\\
    \midrule
    Token Level & T & P & M & K  \\
    \midrule
    supervised & 29.6 & 56.0 & 59.3 & 70.8\\
    \textsc{ULM+GraphInterp} & 40.0 & 60.7 & 61.2 & 77.0\\ 
    \textsc{ULM+GraphFeat}* & 40.1 & \textsc{62.8} & \textsc{63.4} & 78.1\\    
    \bottomrule
  \end{tabular}}
  \caption{ \small{F1 score results on the test set for different categories: T indicates \textsc{Task}, P indicates \textsc{Process}, M is \textsc{Material} and K is Keyword identification (SubTask A). * is transductive model.}}
  \label{tab:summary}
\end{table}

\begin{table*}[t]
  \centering
 {\footnotesize
   \begin{tabular}{p{3cm}p{12cm}}
     \toprule
     Error types &  Annotation  and System Output\\
     \midrule

     Verb phrases & A key requirement in aiming to \blue{[achieve} \cyan{[enantiopure products]$\mathrm{_{\red{\bm{Material}}}}$} \blue{]$\mathrm{_{\bm{Task}}}$} is therefore a means to \blue{[quantitate} \cyan{[the enantiometric excess]$\mathrm{_{\red{\bm{Process}}}}$}\blue{]$\mathrm{_{\bm{Task}}}$}.\\
        \midrule
     General terms & Since the \red{[receptors]}$\mathrm{_{\red{\bm{Material}}}}$  in human biology mostly consist of  \cyan{[chiral molecules]}$\mathrm{_{\cyan{\bm{Material}}}}$, \red{[drug action]}$\mathrm{_{\red{\bm{Process}}}}$ mostly involves a specified enantiometric form. \\
     \midrule
        Falsely predicted adjectives & It has been shown that the most efficient forms of energy transfer between the two occurs when there is a \red{[neighbouring} \cyan{carotenoid species]$\mathrm{_{\bm{Material}}}$}. \\
    \midrule
        Lack of context & Other models use \cyan{[SWEs ]}$\mathrm{_{\blue{\bm{Process}}}^{\red{\bm{Material}}}}$ but focus on the use of multi resolution grids or irregular mesh. \\
     \bottomrule
  \end{tabular}}
  \caption{ \small{Common errors, where \blue{blue} means golden label our system misses, \red{red} means falsely predicted results, and \cyan{green} means correctly predicted spans.}}
  \label{tab:example}
  \end{table*}

 As expected,  the transductive approaches consistently outperform inductive approaches on the test set. With around the same performance on dev set, \textsc{GraphInterp}* seems to generalize better on test set with 1.6\% relative improvement over \textsc{GraphInterp}. We observe higher improvement with \textsc{GraphFeat}* compared to \textsc{GraphInterp}. This is mainly because automatically learning the weight matrix $M$ between neural network scores and graph outputs adds more flexibility compared to tuning an interpolation weight $\alpha$.   
 The performance is further improved by applying data selection through modifying the objective to ULM. The best inductive system is ULM+\textsc{GraphInterp} with 5.6\% relative improvement over pure Self-Training that makes hard decisions, and the best transductive system is ULM+\textsc{GraphFeat}* with 8.6\% relative improvement.

%% file: 07-Analysis.tex
\subsection{Category and Span Analysis}
Table \ref{tab:summary} details the performance of our method on the three categories at the span and token level. We observe significant improvement by using \textsc{ULM+GraphInterp} and \textsc{ULM+GraphFeat} over  best SemEval  and our best supervised system on all three categories   at both  token and span levels. We further observe that systems' performance on  \textsc{Task} classification is much lower than \textsc{Process} and \textsc{Material}. This is in part because \textsc{Task} is much less frequent than the other types. In addition, \textsc{Task} keyphrases often include verb phrases while the other two domains mainly consists of noun phrases. An analysis of confusion patterns show that the most frequent type confusions are between \textsc{Process} and \textsc{Material}. However, we observe that \textsc{ULM+GraphFeat*} can greatly reduce the confusion, with 3.5\% relative improvement of \textsc{Process} and 3.6\% relative improvement of \textsc{Process} over \textsc{ULM+GraphInterp} on token level.



\subsection{Error Analysis}
We provide examples of typical errors that our system makes in Table \ref{tab:example}. As described in the previous subsection, \textsc{task} is the hardest type to identify with our system. Row 1 shows a failure to detect the verb phrase following `to' as part of the \textsc{Task}, but detect `enantiopure products' as \textsc{Material}.  The system prefers to predict \textsc{Process} or \textsc{Material} since those classes have more samples than \textsc{Task}. Row 2 illustrates the problem of identifying general terms as keyphrases due to similar context, such as `receptors' and `drug action'. A third common error involves incorrectly labeling adjectives,  such as `neighbouring' in Row 3, which leads to span errors.  Another common cause of error is insufficient context: in the last example, a larger context  is needed to determine whether `SWE' is a \textsc{Process} or \textsc{Material}.

%% file: 08-Conclusion.tex
\section{Conclusion}
This paper casts the scientific information extraction task as a sequence tagging problem and introduces a hierarchical LSTM-CRF neural tagging model for this task, building on recent results in NER. We introduced a semi-supervised learning algorithm that incorporates graph-based label propagation and confidence-aware data selection. We show the introduction of semi-supervision significantly outperforms the performance of the supervised LSTM-CRF tagging model. We additionally show that external resources  are useful for initializing word embeddings. 
 Both inductive and transductive semi-supervised strategies achieve state-of-the-art performance in SemEval 2017 ScienceIE task. We also conducted a detailed analysis of the system and point out common error cases.
 
 \rev{In our experiments, we observe that including in-domain data only for semi-supervised learning has slightly better performance than using cross-domain data. Reducing the amount of in-domain data hurts performance. Therefore, adding more in-domain unlabeled data may help when combined with selection schemes such as the ULM algorithms proposed here. It would be useful to assess the impact of matched unlabeled data for the physics and material science domain.
 Other future work includes leveraging global context, information of citation network.}

%% file: 09-Acknowledgement.tex
\section{Acknowledgments}

This research was supported by the NSF (IIS 1616112), Allen Institute for AI (66-9175), Allen Distinguished Investigator Award, and gifts from Google, Samsung, and Bloomberg. We thank the anonymous reviewers for their helpful comments.